%% file: template.tex
\RequirePackage{fix-cm}
\documentclass[twocolumn]{svjour3}          %
\smartqed  %
\usepackage{graphicx}

\usepackage{times}
\usepackage{epsfig}
\usepackage{amsmath}
\usepackage{amssymb}
\usepackage{cite}
\usepackage{mathsymb}
\usepackage{textcomp}
\usepackage{verbatim}
\usepackage{url}
\usepackage{multirow}
\usepackage[super]{nth}
\usepackage{colortbl,xspace}

\newcommand{\etc}{\textit{etc.\xspace}}
\newcommand{\etal}{\textit{et al.\ \xspace }}
\newcommand{\eg}{\textit{e.g., \xspace}}
\newcommand{\ie}{\textit{i.e., \xspace}}

\usepackage{hyphenat}

\begin{document}

\title{Deep Recurrent Convolutional Networks for Video-based Person Re-identification: An End-to-End Approach}

\author{Lin Wu   \and Chunhua Shen \and Anton van den Hengel %
}

\institute{L. Wu \and C. Shen \and A. van den Hengel \at
             School of Computer Science, The University of Adelaide, Australia \\
              \email{chunhua.shen@adelaide.edu.au}           %
}

\date{}

\maketitle

\begin{abstract}
In this paper, we present an end-to-end approach to simultaneously learn spatio-temporal features and corresponding similarity metric for video-based person
    re-id\-en\-ti\-fi\-ca\-ti\-on. Given the video sequence of a person, features from each frame that are extracted from all levels of a deep convolutional network can preserve a higher spatial resolution from which we can model finer motion patterns. These low-level visual percepts are leveraged into a variant of recurrent model to characterize the temporal variation between time-steps. Features from all time-steps are then summarized using temporal pooling to produce an overall feature representation for the complete sequence.
The deep convolutional network, recurrent layer, and the temporal pooling are jointly trained to extract comparable hidden-unit representations from input pair of time series to compute their corresponding similarity value.
The proposed framework combines time series modeling and metric learning to jointly learn relevant features and a good similarity measure between time sequences of person.

    Experiments demonstrate that our approach achieves the state-of-the-art performance for video-based person re-i\-de\-n\-ti\-fi\-ca\-ti\-on on iLIDS-VID and PRID 2011, the two primary public datasets for this purpose.
\end{abstract}

\input{introduction.tex}
\input{related.tex}

\input{network.tex}
\input{experiment.tex}

\input{conclusion.tex}

\bibliographystyle{spmpsci}      %
\bibliography{allbib}   %

\end{document}

%% file: introduction.tex
\section{Introduction}\label{sec:intro}

Person re-identification, which aims to recognize an individual over disjoint camera views, is a problem of critical practical importance in video surveillance \cite{Gheissari2006Person,MidLevelFilter,Zheng2011Person,Xiong2014Person,eSDC,Pedagadi2013Local,LADF,Zhao2013SalMatch}.
This problem is very challenging because is is difficult to match pedestrian images captured in different camera views which display large variations in lightings, poses, viewpoints, and cluttered backgrounds. Identifying a person of interest in videos is critical to many surveillance, security and applications such as on-line people tracking in wide-area or off-line people searching in videos.

In this paper, we study the problem of person re\hyp{}identification in the video setting, which occurs when a video of a person as observed in one camera must be matched against a gallery of videos captured by a different non-overlapping camera. The problem of re-identification has been extensively studied for still images, whereas little attention is paid to video based re-identification problem \cite{VideoRanking,VideoPerson,RCNRe-id}.
As a matter of fact, the use of videos for person re-identification exhibits several advantages over still images. In practice, video-based person re-identification provides a more natural method for person recognition in a surveillance system. Videos of pedestrians can easily be captured in a surveillance system, and inevitably contain more information about identity than do a subset of the images therein. Given the availability of sequences of images, temporal priors in relation to a person's motion, such as his/her gait and pose, is captured which may assist in disambiguating difficult cases in recognising an impostor in a different camera. Last but not the least, sequences of images provide more samples of a pedestrian's appearance, where each sample may contain different pose and viewpoint, and thus allows a more reliable appearance based model to be constructed. However, making use of time series also brings about new challenges to re-identification, including the demand of coping with time series of variable length and/or different frame-rates, and the difficulty of seeking a discriminative yet robust feature model given partial or full occlusion within sequences \cite{RCNRe-id}.

\begin{figure*}[t]
\centering
\begin{tabular}{c}
\includegraphics[width=10cm]{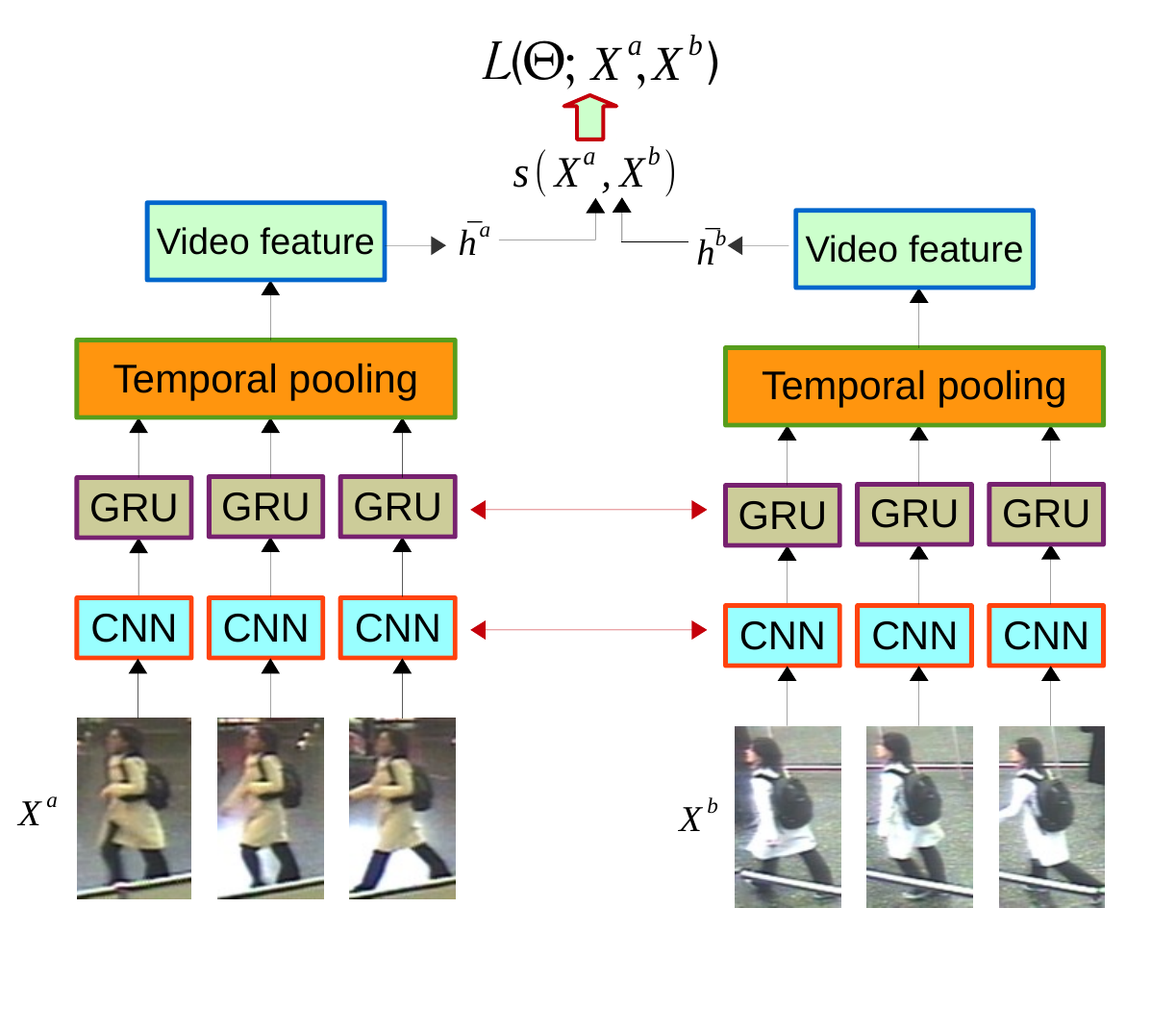}
\end{tabular}
\caption{Given a pair of pedestrian video sequences with variable length (\eg $X^a$ and $X^b$), our deep recurrent convolutional network incorporates nonlinear encoder (three-layer CNNs), recurrent layers (Gated recurrent units), and temporal pooling over time steps to learn video-level features and corresponding similarity value ($s(X^a, X^b)$) in an end-to-end fashion. $L(\Theta; X^a, X^b)$ is the loss function to be minimized and $\Theta$ denotes all network parameters to be learnt. Horizontal red arrows indicate that the two subnetworks share parameterization.}\label{fig:end_to_end}
\end{figure*}

Existing approaches to video-based person re\hyp{}identification \cite{VideoRanking,VideoPerson} are based on low-level features, such as HOG and color histograms, where they either select the most discriminative frame or manually align the temporal sequences, followed by low-level feature extraction. The use of hand-crafted features may not relate well to human appearance, and particularly to how appearance varies between people.

Recent advances in deep learning, especially Recurrent Neural Networks (RNNs) and Long Short-Term Memory (LSTM) models \cite{SeqToSeq,GravesRNN,VideoLSTM,DonahueCVPR2015,DeepVideo,VideoLangModel}, provide some insights as to how to integrate temporal and spatial information. And recurrent convolution networks that leverage both recurrence and convolution have been introduced for learning video representation \cite{VideoLSTM,DonahueCVPR2015,DeepVideo}.
As a consequence, RNN models have been extensively used in action recognition \cite{ConvGRU,RNNHuman,RNNSkeleton,DiffRNN}, where temporal information in an action video can be encoded in conjunction with Convolutional Neural Network (CNN) activations. However, these recurrent convolutional networks customized into actions cannot be directly applicable to person re\hyp{}identification because pedestrians in public yet unconstrained environment exhibit similar walking behaviour without distinctive and semantically categorisable actions unique to different people. Moreover, prior works in action recognition attempt to train the recurrent network by minimizing a ``generative" loss such as negative log-likelihood, which means that such classification model is limited in the case of within-domain to verify new time series of the same person present in training set. Nonetheless, person re-identification imposes the challenge of out-of-domain to verify the time series from pedestrians that are not present in the training set.

\subsection{Our Architecture}
Formally, we consider a supervised learning setting where each training example consists of a pair of sequences $X^a = \{x_1^a, x_2^a, \dots, x_{T_a}^a\}$, $X^b = \{x_1^b, x_2^b, \dots, x_{T_b}^b\}$ in their RGB values along with a class label $y^a$ ($y^b$) for each sequence. At each time-step, images ($x_t^a$ or $x_t^b$) are processed by CNN in which an RNN is applied on all intermediate convolutional layers such that local spatio-temporal patterns can be extracted from convolutional maps that preserve finer spatial resolution of the input video. In the temporal pooling layer, the appearance features produced by the combined CNN and recurrent layer for all time-step are aggregated to generate an overall feature representing the sequence as a whole ($\bar{h}^a$ or $\bar{h}^b$). And the similarity between the time series $X^a$ and $X^b$, \ie $s(X^a,X^b)$, is defined as a weighted inner product between the resulting representations. All parameters of the model ($\Theta$) are learned jointly by minimizing the classification loss on pairs of similar ($y^a=y^b$) and dissimilar time series ($y^a\neq y^b$). The structure of our architecture is shown in Fig. \ref{fig:end_to_end}.

\subsection{Contributions}
The major contributions of this paper is summarized below.
\begin{itemize}
    \item To the best of our knowledge, we deliver the first deep spatio-temporal video representation for person re\hyp{}identification, which is directly learnt from raw sequences of images. This differs significantly from existing methods \cite{VideoRanking,VideoPerson} that are based on hand-crafted features as it automatically learns to extract spatio-temporal features relevant to person re-identification.
\item We present a novel deep recurrent convolutional neural network that combines convolution layer at all levels, recurrent layer, and temporal pooling to produce an improved video representation and learn a good similarity measure between time series.
\item We conduct extensive experiments to demonstrate the state-of-the-art performance achieved by our method for video-based person re-identification.
\end{itemize}

The rest of this paper is structured as follows. In Section \ref{sec:related}, we briefly review related works in terms of video-based person re-identification, and recurrent neural networks. The proposed method is presented in Sections \ref{sec:network} and \ref{sec:network2},
followed by experimental evaluations in Section \ref{sec:exp}. Section \ref{sec:con} concludes this paper.

%% file: related.tex
\section{Related Work}\label{sec:related}

\subsection{Person Re-identification}
The majority of approaches to person re-identification are image-based, which can be generally categorized into two categories. The first pipeline of these methods is appearance based design that aims to extract features that are both discriminative and invariant against dramatic visual changes across views \cite{Farenzena2010Person,MidLevelFilter,Gray2008Viewpoint,eSDC}.
The second stream is metric learning based method which work by extracting features for each image first, and then learning a metric with which the training data have strong inter-class differences and intra-class similarities \cite{Pedagadi2013Local,LADF,LOMOMetric,KISSME,NullSpace-Reid}.

Deep neural network (DNN) approaches have recently been used in person re\hyp{}identification to jointly learn more reliable features and corresponding similarity value for a pair of images \cite{FPNN,JointRe-id,PersonNet,DeepReID,DeepRanking}. In particular, DNNs are used to learn ranking functions based on pairs \cite{DeepReID,PersonNet,FPNN,JointRe-id}, or triplets of images \cite{DeepRanking}. The basic idea is to directly comparing pairs of images and answer the question of whether the two images depict the same person or not. However, these deep architecutres

The use of video in many realistic scenarios indicates that multiple images can be exploited to improve the matching performance. Multi-shot approaches in person
re\hyp{}identification \cite{Farenzena2010Person,Gheissari2006Person,VideoRanking,HumanMatch,SpatioTempPRL}  use multiple images of a person to extract the appearance descriptors to model person appearance.  For these methods, multiple images from a sequence are used to either enhance local image region/patch spatial feature descriptions \cite{Farenzena2010Person,Gheissari2006Person,HumanMatch} or to extract additional appearance information such as temporal change statistics \cite{SpatioTempPRL}.  There are also methods which attempt to select and match
video fragments to maximize cross-view ranking \cite{VideoRanking}. These methods, however, deal with multiple images independently whereas in the video-based person re-identification problem, a video contains more information than independent images, \eg underlying dynamics of a moving person and temporal evolution.

Some previous works on video-based person re\hyp{}identification can be viewed as a process of a two-stage manual sequence alignment and subsequent low-level feature extraction; they extract low-level 3D features (\eg HOG3D \cite{VideoRanking,HOG3D}, color/gradient features over color channels \cite{VideoPerson}) frame by frame through pre-aligned sequences and aggregate these features by encoding them.
However, the temporal sequence nature of videos is not explicitly yet automatically modelled in the their approaches. Moreover, low-level features are still not discriminate enough to distinguish persons and very sensitive to large intra-person visual variations such as illuminations, viewpoint changes, \etc.
More recently, McLaughlin \etal present a recurrent neural network for video-based person re\hyp{}identification \cite{RCNRe-id}.

In this paper, we aim to deliver an end-to-end approach to learn spatio-temporal features and corresponding similarity metric given a pair of time series. Our architecture combines recurrent convolutional network and metric learning by which the convolutional network, and recurrent layer, are jointly trained directly from raw pixels. This can not only act as a feature extractor for video-based re-identification but also produce comparable hidden-unit representations to compute the similarity value for input pair of time series.

\subsection{Spatio-temporal Feature Based Methods}

Spatio-temporal appearance models \cite{Gheissari2006Person,VideoPerson,MatchECCV2012,SpatioTempPRL} treat the video data as a 3D volume and extract local spatio-temporal features. They typically construct spatio-temporal vo-lu-m-e/p-a-t-ch based representations by extending image descriptors such as 3D-SIFT \cite{3DSIFT} and HOG3D  \cite{HOG3D}. In \cite{VideoRanking}, HOG3D is utilized as a spatio-temporal feature representation for person re-identification due to its robustness against cluttered backgrounds and occlusions.
Liu \etal \cite{VideoPerson} proposed a better representation that encodes both the spatial layout of the body parts and the temporal ordering of the action primitives of a walking pedestrian.
However, these methods still construct representations from low-level descriptors including color, texture and gradients.

\subsection{Recurrent Neural Networks in Action Recognition}

By introducing temporal signals into DNNs, some attempts have been made in action/event recognition from video, to understand features occurring over both the spatial and temporal dimensions with recurrent networks \cite{RNNSkeleton,RNNHuman,VideoCNN,ConvGRU}.
For instance, Ballas \etal \cite{ConvGRU} proposed to leverage convolutional units inside an RNN network and they extract temporal pattern from all levels of convolutional maps. However, they focus on learning video representations, and their method relies on pre-trained CNN convolutional maps from ImageNet dataset.

%% file: network.tex
\begin{figure*}[t]
\centering
\begin{tabular}{c}
\includegraphics[width=18cm]{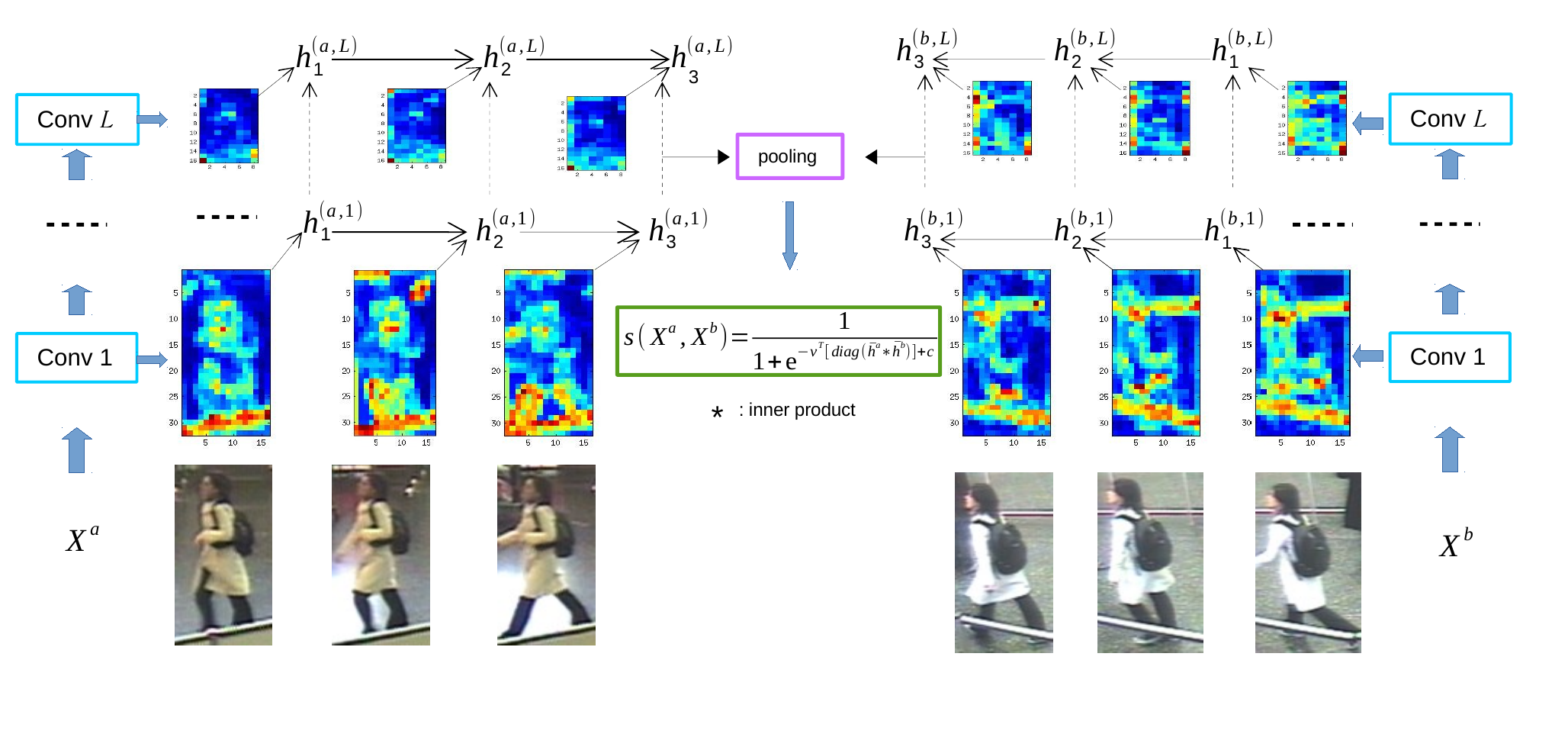}
\end{tabular}
\caption{Graphical illustraion of our architecture. Given a pair of pedestrian video sequences (\eg $X^a$ and $X^b$), our model is to jointly train convolution network, and recurrent layer in order to learn spatio-temporal features and corresponding similarity value simultaneously. In particular, at each time step $t$, convolutional resolutions at all levels are leveraged and modeled by deep recurrent convolutional network to estimate hidden representation \ie $h_t^{(a, l)}$ ($h_t^{(b, l)}$). The similarity is defined by inner product of two hidden-unit activations $\bar{h}^a$ and $\bar{h}^b$(after pooling).}\label{fig:DeepRCN}
\end{figure*}

\section{Gated Recurrent Unit Networks}\label{sec:network}

In this section, we review a particular type of RNN model, Gated Recurrent Unit (GRU) networks. Following the concepts and notations defined in GRU networks, in Section \ref{sec:network} we develop a deep recurrent convolutional neural network in which a GRU is a recurrent unit to adaptively capture dependencies of different time scales.

Naturally suited for variable-length inputs like video sequences, an RNN model defines a recurrent hidden state whose activation at each time step is dependent on that of the previous time. Given a sequence $X=\{x_1,x_2,\dots,x_T\}$, the hidden state at time $t \in \{1,\dots,T\}$ is defined as $h_t=sigmoid (W x_t + U h_{t-1})$. Standard RNNs are hard to train due to the well-known vanishing or exploding gradient problems \cite{DifficultRNN1994,DifficultRNN2012}. To address the vanishing gradient problem, the Long Short Term Memory (LSTM) architecture was proposed for learning long range dependencies through its use of memory cell units that can store/access the information across lengthy input sequences \cite{lstm1997,GravesRNN}.
However, LSTMs typically require large training datasets to achieve good generalization due to their vast numbers of parameters.  As an alternative to LSTM, the GRU architecture was proposed by Cho \etal \cite{GRU2014}, and after extensive empirical analysis, GRUs were found to have similar performance to LSTMs but with fewer parameters \cite{EmpiricalRNN,EmpiricalGRU}.
In this paper, we implement GRU as the recurrent units due to its superior performance in many tasks such as music/speech signal modeling \cite{EmpiricalGRU} and fewer parameters as opposed to LSTMs.

The GRU is formulated as follows:

\begin{equation}\label{eq:fully_GRU}
\begin{split}
& z_t = \sigma(W_z x_t + U_z h_{t-1}),\\
& r_t = \sigma(W_r x_t + U_r h_{t-1}),\\
& \hat{h}_t = tanh(W x_t + U(r_t \odot h_{t-1})),\\
& h_t = (1-z_t)h_{t-1} + z_t \hat{h}_t,
\end{split}
\end{equation}
where $\odot$ is an element-wise multiplication. $z_t$ is an update gate that determines the degree to which the unit updates its activation. $r_t$ is a reset gate, and $\sigma$ is the sigmoid function. The candicate activation $\hat{h}_t$ is computed similarly to that of traditional recurrent unit. When $r_t$ is close to zero, the reset gate make the unit act as reading the first symbol of an input sequence and forgets the previously computed state.

The formulation of GRU in Eq\eqref{eq:fully_GRU} is a \textit{fully-connected} GRU (FC-GRU) where the input, and states are all 1D vectors.
In this paper,  we aim to model the spatial-temporal relationships in video sequence by taking the inputs of convolutional maps, thus we extend the idea of fully-connected GRUs to convolutional GRUs (ConvGRU) which has convolutional structures in both the input-to-hidden and hidden-to-hidden transitions. We will present the idea of ConvGRU in Section \ref{sec:deep_rcn}.

\section{Deep Recurrent Convolutional Networks for Video-based Person Re-identification}\label{sec:network2}

A graphical depiction of our deep convolutional-recurrent architecture is shown in Fig.\ref{fig:DeepRCN}. At each time step $t$, frames $x_t^a$ and $x_t^b$ of sequence $X^a$ and $X^b$ passes through the encoder, the recurrent layers, and the pooling layer, producing the output of a similarity value.
Similar to a Siamese network \cite{Siamese},
our architecture employs two encoder-recurrent networks that share their parameters in order to extract comparable hidden-unit representations from time series inputs. The encoder to extract spatial resolutions is essentially a three layer convolutional network with architecture similar to \cite{PersonNet}.

\subsection{Convolutional Layer}

The encoder in our architecture corresponds to three-layer CNNs, which can transfer the input data to a representation where learning of pedestrian motion dynamics is easy \cite{RNNHuman}.

During training, the input to our network is a mini-batch containing pairs of fixed-size 160$\times$60 RGB images. The images are passed through four convolutional layers, where we use filters with a very small receptive filed: 3$\times$3. The convolution stride is fixed to 1 pixel. Spatial pooling is carried out by three max-pooling layers. Max-pooling is performed over a 2$\times$2 pixel window, with stride 2. After a stack of convolution layers, we have two fully-connected layers where the first two have 4096 dimension and the third is 512, which are then fed into the last hash layers to generate a compact binary code. We show details of layers in CNNs in Table \ref{tab:layer}.

\begin{table}[t]
  \centering
  \caption{Layer parameters of convolutional neural networks in encoder. The output dimension is given by height$\times$width$\times$width. FS: filter size for convolutions. Layer types: C: convolution, MP: max-pooling.}  \label{tab:layer}
  {
  \begin{tabular}{c|c|c|c|c}
  \hline
\hline
    Name  & Type  &  Output Dim & FS  & Sride \\
  \hline\hline
   Conv0  & C & $157\times57\times32$  & 3$\times3$ & 1 \\
   Pool0  & MP &  $79\times 29\times 32$ & 2$\times2$ & 2 \\
   Conv1  & C & $76\times 26\times 32$ & 3$\times$3 & 1 \\
   Pool1  & MP & $38\times 13\times 32$ & 2$\times$2 & 2 \\
   Conv2  & C &  $35\times 10\times 32$ &  3$\times$3  & 1 \\
   Pool2  & C & $18\times 5\times 32$ & 3$\times$3 &  2\\
   Conv3  & C & $15\times 2\times 32$ & 3$\times$3 &  1\\
\hline
  \hline
  \end{tabular}
  }
\end{table}

\subsection{Recurrent Layer}\label{sec:deep_rcn}

On top of the encoder part, we apply $L$ RNNs on each convolutional map. The $L$ RNNs can be parameterized as $\phi^1$, $\phi^2$, $\dots$, $\phi^L$, such that $h_t^l=\phi^l(x_t^l, h_{t-1})^l$.
The output feature representations $\bar{h}$ are obtained by average pooling the hidden unit activations over final time steps $h_T^1,h_T^2,\dots,h_T^L$. In this paper, the recurrent function $\phi^l$ is implemented by using Gated Recurrent Units \cite{GRU2014} where GRUs were originally introduced to model input to hidden state and hidden to hidden state transitions using fully connected units. However, the output convolutional maps of the encoder subordinate to recurrent layer are 3D tensors (spatial dimensions and input channels) for which directly applying a FC-GRU can lead to a massive number of parameters and too much redundancy for these spatial data. Let $H_1$ and $H_2$ and $C_x$ be the input convolutional map spatial dimension and number of channels. A FC-GRU would require the input-to-hidden parameters $W_z^l$, $W_r^l$ and $W^l$ of size $H_1 \times H_2 \times C_x \times C_h$ where $C_h$ is the dimensionality of the GRU hidden representation.

Inspired by Convolutional LSTMs \cite{ConvLSTM}, we employ convolutional GRU (ConvGRU) to capture the motion dynamics  involved over time series. The ConvGRUs are first introduced by Ballas \etal \cite{ConvGRU} to take advantage of the underlying structure of convolutional maps.
The rationale of using ConvGRUs is based on two observations: 1) convolutional maps extracted from images represent strong \textbf{local correlation} among responses, and these local correlations may repeat over different spatial locations due to the shared weights of convolutional filters (a.k.a. kernels) across the image spatial domain; 2) videos are found to have strong \textbf{motion smoothness} on temporal variation over time, \eg motion associated with a particular patch in successive frames will appear and be restricted within a local spatial neighborhood.

\begin{figure}[t]
\centering
\includegraphics[height=4cm]{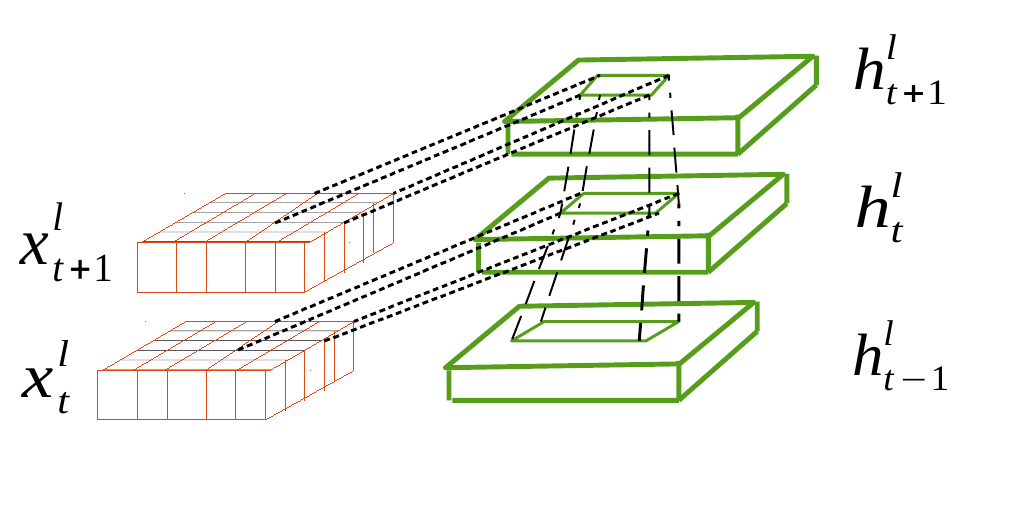}
\caption{The structure of the convolutional GRU. $x_t^l$ and $x_{t+1}^l$ have dimension $a\times b\times M$, corresponding to the frame-level CNN features of layer $l$. To encode the spatial information into hidden state $h_{t+1}^l$, the state in the grid is determined by the input ($x_t^l$) and the past state of its local neighbors ($h_{t-1}^l$), as implemented by a convolutional operation. }
\label{fig:ConvGRU}
\end{figure}

In this paper, we propose to reduce the computational complexity of the deep recurrent neural networks by further sparsifying their hidden unit connection structures. Specifically, we embed the priors of locality and motion smoothness in our model structure and replace the fully-connected units in GRU with convolutional filters (kernels). Thus, recurrent units are encouraged to have sparse connectivity (h\-i\-dd\-e\-n-to-hi\-dd\-e\-n) and share their parameters across different input spatial locations (input-to-hidden), and we can formulate a ConvGRU as:

\begin{equation}\label{eq:ConvGRU}
\begin{split}
& z_t^l = \sigma(W_z^l * x_t^l + U_z^l * h_{t-1}^l),\\
& r_t^l = \sigma(W_r^l * x_t^l + U_r^l * h_{t-1}^l),\\
& \hat{h}_t^l = tanh(W^l * x_t^l + U^l * (r_t^l \odot h_{t-1}^l )),\\
& h_t^l = (1-z_t^l)h_{t-1}^l + z_t^l \hat{h}_t^l,
\end{split}
\end{equation}
where * denotes a convolution operation.  In Eq \eqref{eq:ConvGRU}, input-to-hidden parameters $W_z^l$, $W_r^l$, $W^l$, and hidden-to-hidden parameters $U_z^l$, $U_r^l$, $U^l$ are 2D convolutional kernels with size of $k_1\times k_2$ where $k_1\times k_2$ is much smaller than original convolutional map size $H_1 \times H_2$. The input and hidden representation can be imagined as vectors standing on spatial grid, and the ConvGRU determines the future hidden representation in the grid by the inputs and past hidden units of its local neighbors. The resulting hidden recurrent representation $h_t^l = \{h_t^l(i,j)\}$ can preserve the spatial structure where $h_t^l(i,j)$ is a feature vector defined at the grid location $(i,j)$. To ensure the spatial size of the hidden representation remains the same over time, zero-padding is needed before applying the convolution operation.
The structure of a convolutional GRU is illustrated in Fig.\ref{fig:ConvGRU}.

In fact, the hidden recurrent representation of moving objects is able to capture faster motiong moving with a larger transitional kernel size while one with a smaller kernel can capture slower motion. Recall that FC-GRU model inputs and hidden units as vectors, which can be viewed as 3D tensors with the last dimension being 1.
In this sense, FC-GRU becomes a special case of ConvGRU with all features being a single vector.

\subsection{Complexity}

By virtue of convolution operation, input-to-hidden parameters $W_z^l$, $W_r^l$, $W^l$ have the size of $k_1\times k_2 \times C_x \times C_h$. The activation gate $z^l_t(i,j)$, the reset gate $r^l_t(i,j)$, and the candidate hidden representation $\hat{h}_t^l (i,j)$ are defined based on a local neighborhood of size $k_1\times k_2$ at the location $(i,j)$ on both the input data $x_t^l$ and the previous hidden units $h^l_{t-1}$. A ConvGRU layer conducts 2D convolutional operation 6 times at each time step \ie 2 convolutions per GRU gate and another 2 in computing $\hat{h}^l_t$. Assume the input-to-hidden and hidden-to-hidden transition parameters have the same kernel size while preserving the input dimensions, the ConvGRU requires $O(3 T H_1 H_2 k_1 k_2 (C_x C_h + C_h^2))$ computations where $T$ is the length of a time sequence. It is apparent that ConvGRU saves computations substantially compared with a FC-GRU which requires  $O(3 T H_1^2 H_2^2 (C_x C_h + C_h^2))$ multiplications. In memory, a ConvGRU only needs to store parameters for 6 convolutional kernels, yielding \[
    O(3 k_1 k_2 (C_x C_h + C_h^2))
    \]
parameters.

\subsection{Deep Recurrent Convolutional Networks}

To leverage $L$ ConvGRUs which are associated with $L$ convolutional maps at each time step, we stack these ConvGRUs by preconditioning each ConvGRU on the output of the previous ConvGRU at the current time step:
\[
    h^l_t =\phi^l (h_{t-1}^l, h_t^{l-1}, x_t^l).
    \]

Specifically, at time step $t$, the lower-level recurrent hidden representation $h_t^{l-1}$ is considered as  an extra input to the current convolutional GRU units:

\begin{equation}\label{eq:DeepGRU}
\begin{split}
& z_t^l = \sigma(W_z^l * x_t^l + W^l_{z^l} * h_t^{l-1} +  U_z^l * h_{t-1}^l),\\
& r_t^l = \sigma(W_r^l * x_t^l + W^l_{r^l} * h_t^{l-1} + U_r^l * h_{t-1}^l),\\
& \hat{h}_t^l = tanh(W^l * x_t^l + U^l * (r_t^l \odot h_{t-1}^l )),\\
& h_t^l = (1-z_t^l)h_{t-1}^l + z_t^l \hat{h}_t^l.
\end{split}
\end{equation}

The advantage of adding this cross-layer hidden connections is to enable the model to leverage representations with different resolutions since these low-level spatial resolutions certainly encode informative motion patterns. In particular, the receptive field associated with each $h_t^l (i,j)$  increases in previous hidden states $h^l_{t-1}$, $h^l_{t-2}$,\dots, $h^l_{1}$ if we go back along the time line. Finally, an average pooling is introduced on the top of the network to combine the features at all convolutional levels to generate overall feature appearance for the complete sequence:
\begin{equation}
\bar{h} = \frac{1}{L}  \sum_{l=1}^L h_T^l.
\end{equation}

\subsection{Parameter Learning}

Suppose we are given two time series $X^{a}=\{x_1^a,x_2^a,\dots,x_{T_a}^a\}$ and $X^{b}=\{x_1^b,x_2^b,\dots,x_{T_b}^b\}$. The hidden unit representations $\bar{h}^a$ and $\bar{h}^b$ computed from the two subnetworks for the corresponding input time series can be combined to compute the prediction for the similarity of the two time series. Thus, we define the similarity of the two sequences as:
\begin{equation}\label{eq:similarity}
s(X^a,X^b) = \frac{1}{1 + e^{-v^T [diag(\bar{h}^a * \bar{h}^b)] + c}},
\end{equation}
where the element-wise inner product between the hidden representations is computed \ie $\bar{h}^a * \bar{h}^b$, and the output is a weighted sum of the resulting product ($v$ is the weighted vector). $c$ is a constant value. Herein, the similarity between two time series are defined as a weighted inner product between the latent representations $\bar{h}^a$ and $\bar{h}^b$. Thus, time series modeling and metric learning can be combined to learn a good similarity value. Indeed, such similarity measure between hidden-units activations have previously been used as a part of attention mechanisms in speech recognition \cite{SiameseRNN,SpeechRNN}, and handwriting generation \cite{GravesRNN}.

Let $Z$ denote a training set containing two sets of pairs of time series: a set with pairs of similar time series $S$ and a set with pairs of dissimilar time series $D$. We aim to learn all parameters $\Theta =\{ W_z^l, W_r^l, W^l, U_z^l, U_r^l, U^l, v, c\}$ in our network jointly by minimizing the binary cross-entropy prediction loss. This is equivalent to maximizing the conditional log-likelihood of the training data:
\begin{multline}
    \label{eq:loss}
    \mathcal{L}(\Theta; Z) =
    \\
    -\left[ \sum_{(a,b)\in S} \log s(X^a, X^b)
    + \sum_{(a,b)\in D}  \log (1-s(X^a, X^b))
       \right],
\end{multline}
where $a$ and $b$ denote the indices of two time sequences in training pairs.

The loss function is maximized and back-propagated through both recurrent networks and stacked convolutional layers (the weights of two subnetworks are shared) using a variant of the back-propagation through time algorithm with gradient clipping between -5 and 5 \cite{AdvanceRNN}.
The sets $S$ and $D$ are defined using class labels for each time sequence: $S=\{ (a,b): y^a = y^b\}$ and $D=\{(a,b): y^a \neq y^b\}$ where $y^a$ and $y^b$ are class labels for time sequences $X^a$ and $X^b$. In the case of person re-identification, each person can be regarded as a class, and thus training class label can be assigned on each sequence accordingly. In contrast existing classification deep models in person re-identification \cite{JointRe-id,FPNN,PersonNet}, our loss function allows our architecture to be applied on objects from unknown classes.  For instance, a classification deep model trained on a collection of pedestrian images can be used within-domain to verify new images of the same people. However, our network can be trained and applied on out-of-domain to verify the new images from pedestrians that were not present in training set. Thus, this loss function is more suitable to person re-identification where the underlying assumption is that inter-person variations have been modeled well.

%% file: experiment.tex
\section{Experiments}\label{sec:exp}

In this section , we present an empirical evaluation of the proposed deep RCN architecture. Experiments are conducted on two benchmark video sequences for person re-identification.

\subsection{Experimental Setting}

\paragraph{Datasets}

We validate our method and compare to state-of-the-art approaches on two video-based image sequence datasets designed for video-based  person re-identification: the iLIDS-VID dataset \cite{VideoRanking} and the PRID 2011 dataset \cite{PRID2011}.

\begin{figure}[t]
\centering
\includegraphics[height=3cm,width=6cm]{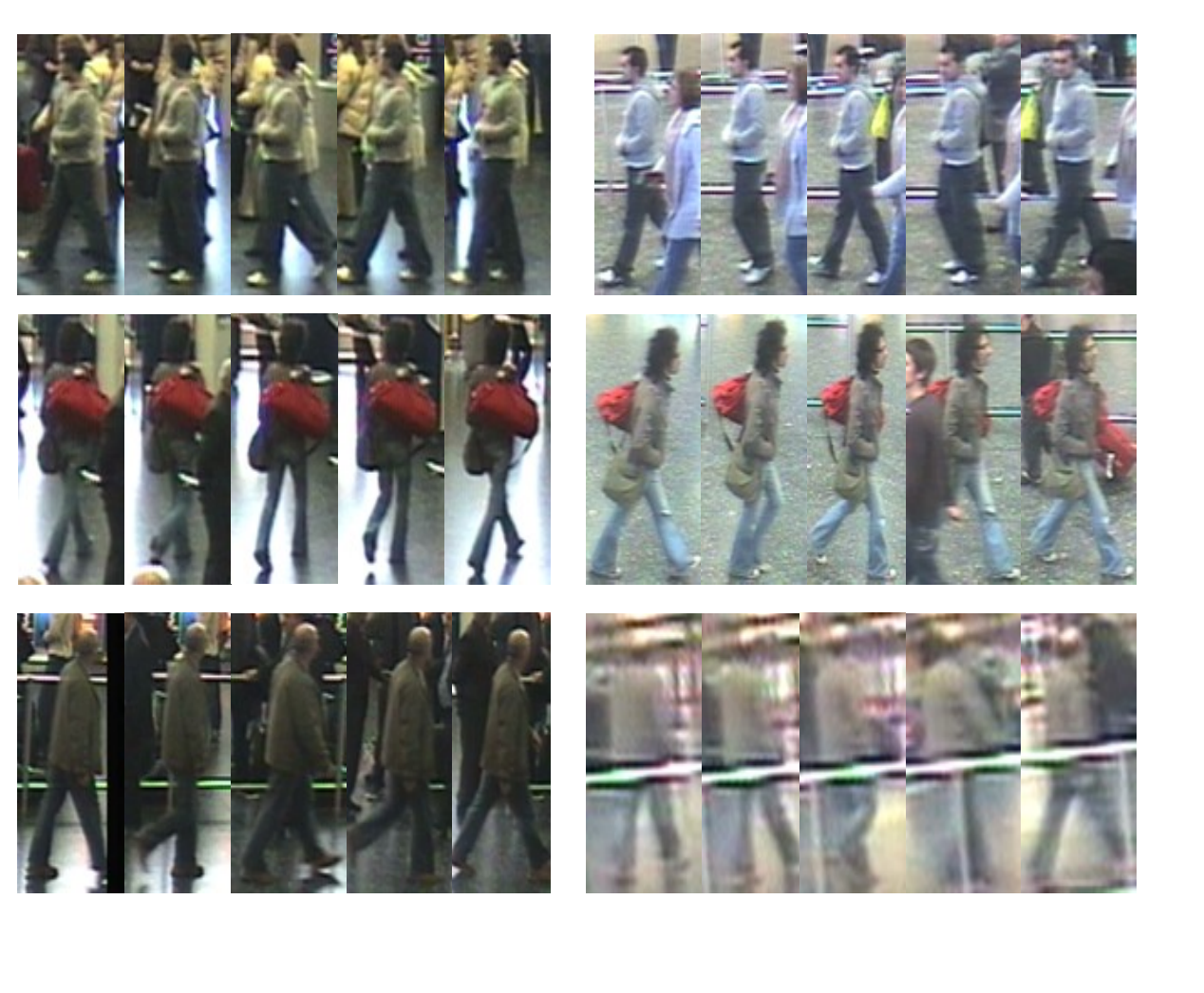}
\includegraphics[height=3cm,width=6cm]{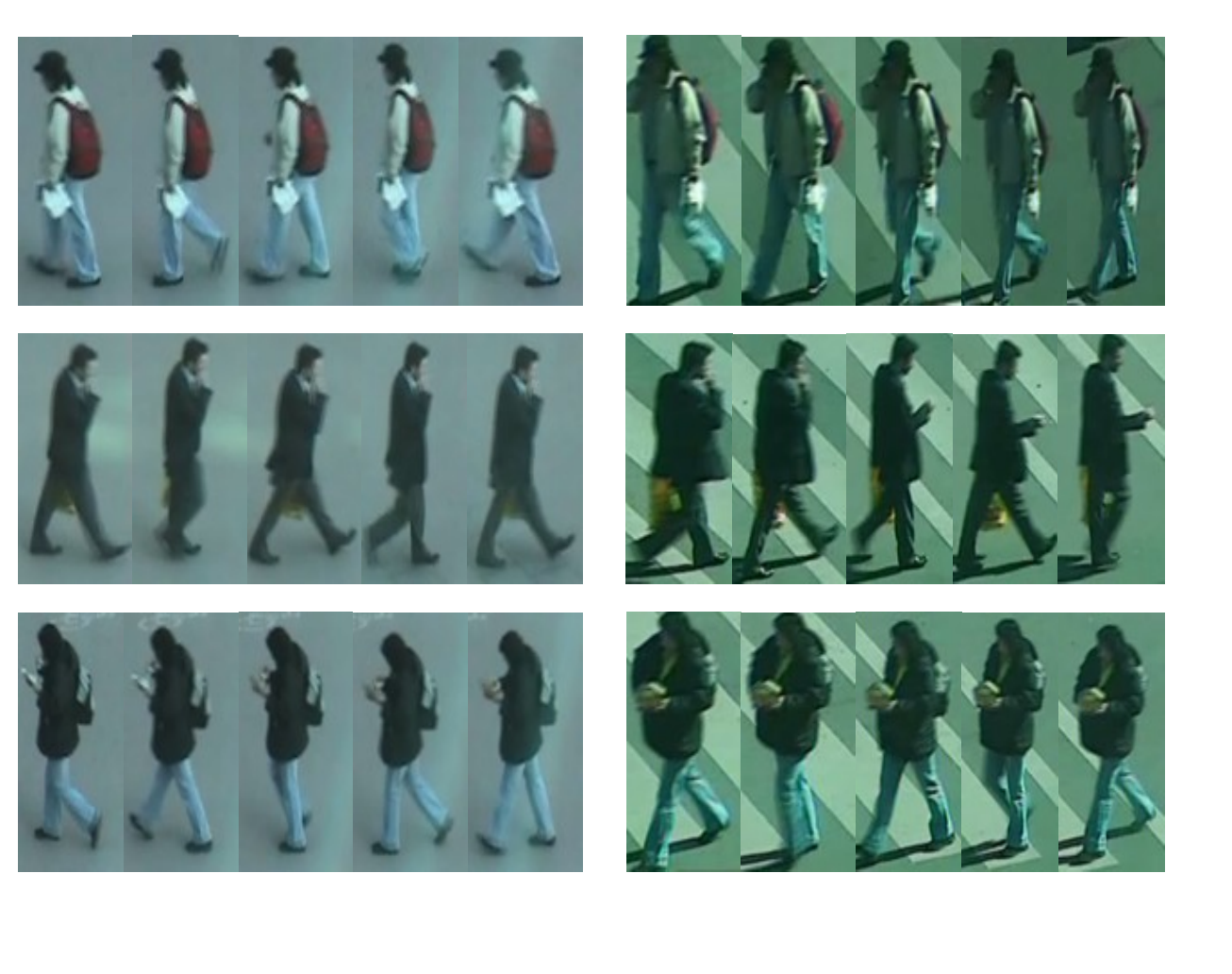}\\
(a) iLIDS-VID (b) PRID2011
\caption{Example pairs of image sequence of the same pedestrian in different camera views from the two datasets.}
\label{fig:examples}
\end{figure}

\begin{itemize}
\item The iLIDS-VID dataset consists of 600 image sequences for 300 randomly sampled people, which was created based on two non-overlapping camera views from the i-LIDS multiple camera tracking scenario. The sequences are of varying length, ranging from 23 to 192 images, with an average of 73. This dataset is very challenging due to variations in lighting and viewpoint caused by cross-camera views, similar appearances among people, and cluttered backgrounds. (See selected examples in Fig. \ref{fig:examples}(a).)
\item The PRID 2011 dataset includes 400 image sequences for 200 persons from two adjacent camera views. Each sequence is between 5 and 675 frames, with an average of 100. Compared with iLIDS-VID, this dataset was captured in uncrowded outdoor scenes with rare occlusions and clean background. However, the dataset has obvious color changes and shadows in one of the views. (See selected examples in Fig.\ref{fig:examples}(b).)
\end{itemize}

\subsection{Model Training}

We implemented our architecture in Python with the help of Theano \cite{Theano,TheanoNew}. All experiments are run on a single PC with a single NVIDIA GTX 980 GPU. During training, sub-sequences of $T=20$ consecutive frames are used for computational purpose, where a different subset of 20 frames are randomly selected from video sequence at each epoch. In test, we regard the first camera as the probe while the second camera as the gallery \cite{VideoRanking}.
We did self data augmentation in training by cropping and mirroring all frames for a given sequence. In testing, data augmentation is also applied into the probe and the gallery sequences, and the similarity scores between sequences are computed and averaged over all augmentation conditions \cite{AugmentationREID,ImproveCNN}.

In our deep GRU-RCN, all input-to-hidden and hidden-to-hidden kernels are of size $5\times 5$, \ie $k_1=k_2=5$. We apply zero-padded $5\times 5$ convolutions on each ConvGRU to preserve the spatial dimension. The stacked GRU-RCN consists of 3 convolutional RNNs with 128, 256, 256 channels, respectively. Recall that each ConvGRU is preconditioned on the hidden\hyp{}representation of the previous convolution maps. Max-pooling is applied on hidden-representations between the recurrent layers for the compatability of the spatial dimensions. The hidden\hyp{}representation at the last time step is pooled and then given as the overall representation for the complete sequence.

In our experiments, parameters of the model are initialized by sampling them from an uniform distribution within an interval [-1, 1]. Training of the model is performed using a RMSProp \cite{RMSProp} stochastic gradient decent procedure using mini-batches of 10 pairs of time series. To prevent the gradients from exploding, we clip all gradients to lie in the interval [-5, 5] \cite{AdvanceRNN}. A dropout ratio of 0.7 is applied on the hidden-unit activations. As suggested in Section \ref{sec:network}, we define similar video sequence to be those from the same person and dissimilar ones to be those from different persons.

\subsection{Competitors and Evaluation Metric} We compare the performance of our deep RCN with that of two variants of our model, and with a competitors with fully-connected GRU. The two variants of our model are: (1) An RCN architecture with three ConvGRUs independently on each convolution map, denoted as \textit{RCN-Ind}. It uses the same base architecture as the deep GRU-RCN. After the convolution operation, we obtain 3 hidden-representation for each time step, and we apply average pooling on the
hidden\hyp{}representation of the last time step to reduce their spatial dimension to $1\times 1$. (2) We modify the network configurations with the input-to-state and state-to-state kernels of size $9\times 9$ and $1\times 1$. The number of parameters with this modified network is close to the original one.

The competitor with FC-GRU are based on VGG 16 model \cite{VGG}, pretrained on ImageNet \cite{ImageNet}, and fine-tune it on the single shot VIPeR re-identification dataset \cite{Gray2007}. We extract features from fully-connected layer $fc7$ (which can be viewed as a feature map with $1\times1$ spatial dimension), which are given as the inputs to our deep network and act as FC-GRU.

\paragraph{Evaluation Metric}
In our experiments, each dataset is randomly split into 50\% of persons for training and 50\% of persons for testing. The performance is evaluated by the average Cumulative Matching Characteristics (CMC) curves after 10 trials with different train/test split.

\subsection{Results}

\subsubsection{Comparison with Architecture Variants}
Our experiments show that deep RCN with ConvGRUs perform consistently better than VGG-16 FC-GRU. Also our model with stacked RCN can give better results than RCN-Ind although the improvement is not significant.

\begin{table*}[t]
\caption{Evaluation on architecture variants.}\label{tab:com_arch}
\centering
\begin{tabular}{l|c|c|c|c|c|c|c|c}
\hline\hline
Dataset & \multicolumn{4}{c|}{iLIDS-VID} & \multicolumn{4}{c}{PRID2011}\\
\cline{1-9}
Rank @ R & R = 1 & R = 5 & R = 10 & R = 20 & R = 1 & R = 5 & R = 10 & R = 20\\
\hline
RCN-Ind & 41.4 & 66.9 & 84.7  & 91.0 & 48.2  & 77.0 & 87.9 & 93.0\\
VGG 16 + FC-GRU &  38.6 & 62.5 & 81.3 & 88.5 & 42.2 & 71.8 & 80.2 & 89.7\\
Deep RCN ($9\times 9$, $1\times 1$)  & 40.1  & 66.7  & 86.1 & 90.5 & 47.1 & 74.2 & 87.5 & 92.1\\
Deep RCN ($5\times 5$, $5\times 5$) & \textbf{42.6} & \textbf{70.2} & \textbf{86.4} & \textbf{92.3} & \textbf{49.8} & \textbf{77.4} & \textbf{90.7} & \textbf{94.6}\\
\hline
\hline
\end{tabular}
\end{table*}

\subsubsection{Study on Pooling Layer}

A standard way to achieve a video-level representation where the local descriptor relies on recurrency at frame-level is to first apply normalization on frame descriptors and then average pooling of the normalised descriptors over time to obtain the final representation. Thus, the average pooling over temporal frames is defined as $\textbf{x}_{video}=\frac{1}{N}\sum_{i=1}^N \textbf{x}_i$, where $\textbf{x}_i$ is the frame-level descriptor output from the RCN and $N$ is the total number of frames.
A common alternative is to apply max pooling  similarly. Fisher vector is \dots

We thus compare the performance using the other two pooling methods. Results are given in Table \ref{tab:com_pooling}. It can be seen that Fisher vector encoding performs slightly better than max- and averaging pooling. This is mainly because Fisher vectors use higher order statistics to model feature elements in a generative process. Due to the high computational cost of updating Fisher vector parameters, we use average pooling as the temporal pooling throughout the experiments.

\begin{table*}[t]
\caption{Evaluation on different feature pooling schemes on top of convolutional GRUs.}\label{tab:com_pooling}
\centering
\begin{tabular}{l|c|c|c|c|c|c|c|c}
\hline\hline
Dataset & \multicolumn{4}{c|}{iLIDS-VID} & \multicolumn{4}{c}{PRID2011}\\
\cline{1-9}
Rank @ R & R = 1 & R = 5 & R = 10 & R = 20 & R = 1 & R = 5 & R = 10 & R = 20\\
\hline
Max pooling & 39.4 & 68.6 & 83.8  &  90.4 & 42.4  & 72.8 & 89.0 & 90.2\\
Average pooling & 42.6 & 70.2 & 86.4 & 92.3 & \textbf{49.8} & 77.4 & 90.7 & \textbf{94.6}\\
Fisher vector pooling & \textbf{43.2} & \textbf{70.7} & \textbf{88.2}  & \textbf{93.5} & 49.2  & \textbf{77.7} & \textbf{91.4} & 94.1\\\hline
\hline
\end{tabular}
\end{table*}

\subsubsection{Comparison with Other Representations}

\begin{table*}[t]
\caption{Comparison with different feature representations.}\label{tab:com_feature}
\centering
\begin{tabular}{l|c|c|c|c|c|c|c|c}
\hline\hline
Dataset & \multicolumn{4}{c|}{iLIDS-VID} & \multicolumn{4}{c}{PRID2011}\\
\cline{1-9}
Rank @ R & R = 1 & R = 5 & R = 10 & R = 20 & R = 1 & R = 5 & R = 10 & R = 20\\
\hline
HOG3D \cite{HOG3D}  & 8.3 & 28.7 & 38.3 & 60.7 & 20.7 & 44.5 & 57.1 & 76.8\\
FV2D  \cite{LocalFV} & 18.2 & 35.6 & 49.2 & 63.8 & 33.6 & 64.0 & 76.3 & 86.0\\
FV3D \cite{VideoPerson} & 25.3 & 54.0 & 68.3 & 87.3 & 38.7 & 71.0 & 80.6 & 90.3\\
STFV3D \cite{VideoPerson} & 37.0 & 64.3 & 77.0 & 86.9 & 42.1 & 71.9 & 84.4 & 91.6\\
Deep RCN & \textbf{42.6} & \textbf{70.2} & \textbf{86.4} & \textbf{92.3} & \textbf{49.8} & \textbf{77.4} & \textbf{90.7} & \textbf{94.6}\\
\hline
\hline
\end{tabular}
\end{table*}

In this section, we compare our feature representation with  four competing representations for person re-identification.
\begin{itemize}
\item \textbf{FV2D} is a multi-shot approach \cite{LocalFV} which treats the video sequence as multiple independent images and uses Fisher vectors as features.

\item \textbf{HOG3D} extracts 3D HOG features \cite{HOG3D} from volumes of video data \cite{VideoRanking}. Specifically, after extracting a walk cycle by computing local maxima/minima of the FEP signal, video fragments are further divided into $2\times 5$ (spatial) $\times$ 2 (temporal) cells with 50\% overlap. A spatio-temporal gradient histogram is computed for each cell which is concatenated to form the HOG3D descriptor.

\item \textbf{FV3D} is similar to HOG3D where a local histogram of gradients is extracted from divided regular grids on the volume. However, we encode these local HOG features with Fisher vectors instead of simply concatenating them.

\item \textbf{STFV3D} is a low-level feature-based Fisher vector learning and extraction method which is applied to spatially and temporally aligned video fragments \cite{VideoPerson}. STFV3D proceeds as follows: 1) temporal segments are obtained separately by extracting walk cycles \cite{VideoRanking}, and spatial alignment is implemented by detecting spatial bounding boxes corresponding to six human body parts; 2) Fisher vectors are constructed from low-level feature descriptors on those body-action parts.

\end{itemize}

Experimental results are shown in Table \ref{tab:com_feature}. From the results, we can observe that our deep representation outperforms consistently over other representations. More specifically, HOG3D is inferior to Fisher vectors based features since Fisher vectors encode local descriptors in a higher order and suitable for person re-identification problem. It is not a surprise to see our features are superior to STFV3D because our deep RCNs work well in reconstructing spatio-temporal patterns.

\subsection{Comparison with State-of-the-art Approaches}\label{ssec:state-of-the-art}

In this section, we evaluate and compare our method with state-of-the-art video-based person re-identification approaches. To enable a fair comparison, we combine two supervised distance metric learning methods: Local Fisher Discriminant Analysis (LFDA \cite{Pedagadi2013Local}) and KISSME \cite{KISSME}. PCA is first applied to reduce the dimension of our original representation, and we set the reduced dimension as 128 in the implementation. In this experiment, competitors include
\begin{itemize}
\item Gait Energy Image + Rank SVM, denoted as GEI+RSVM \cite{GaitRank};
\item HOG3D + Discriminative Video Ranking (DVR) \cite{VideoRanking};
\item Color + LFDA \cite{Pedagadi2013Local};
\item STFV3D+LFDA/KISSME \cite{VideoPerson}.
\end{itemize}

\begin{figure}[t]
\centering
\includegraphics[height=5cm]{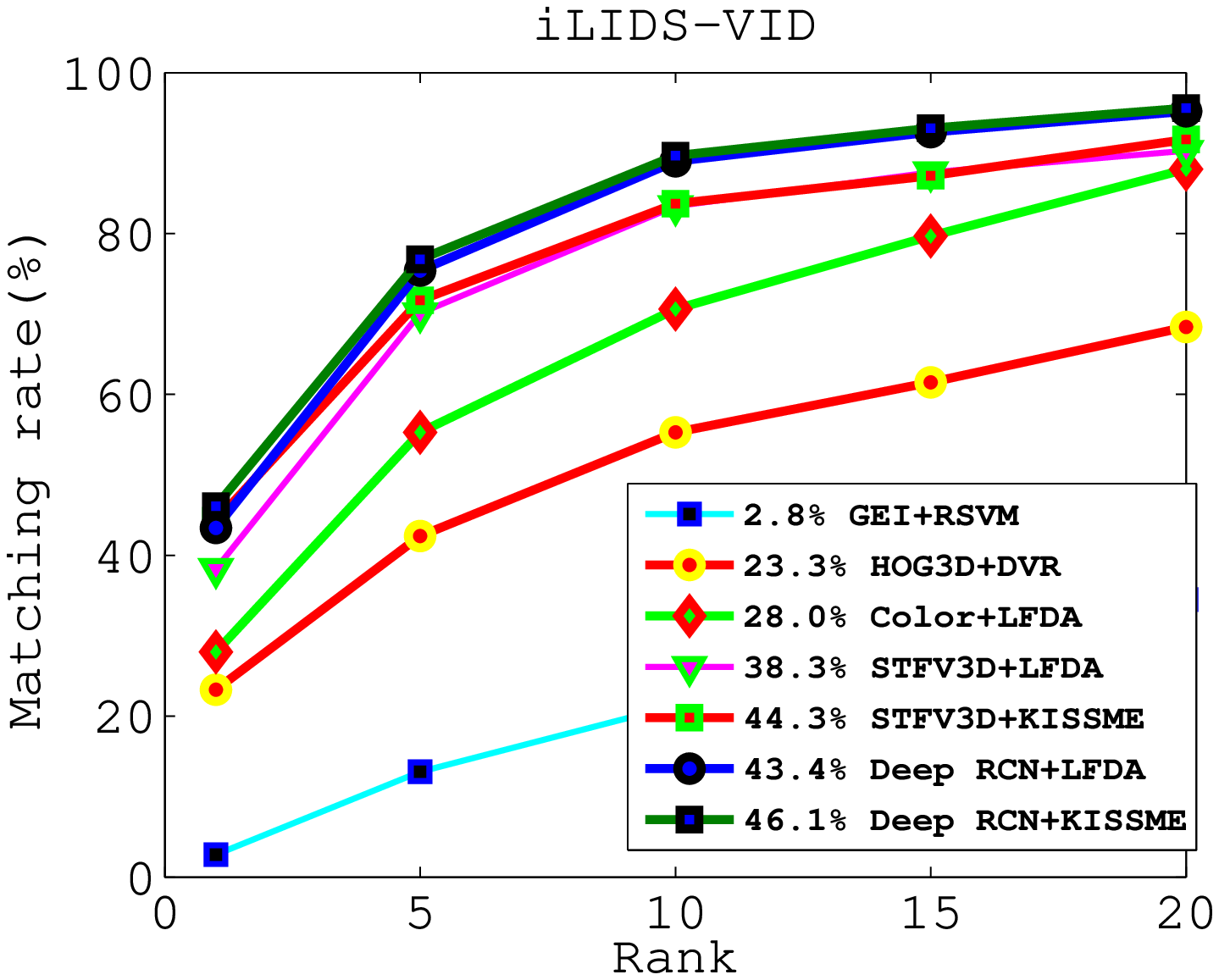}\\
(a) iLIDS-VID\\
\includegraphics[height=5cm]{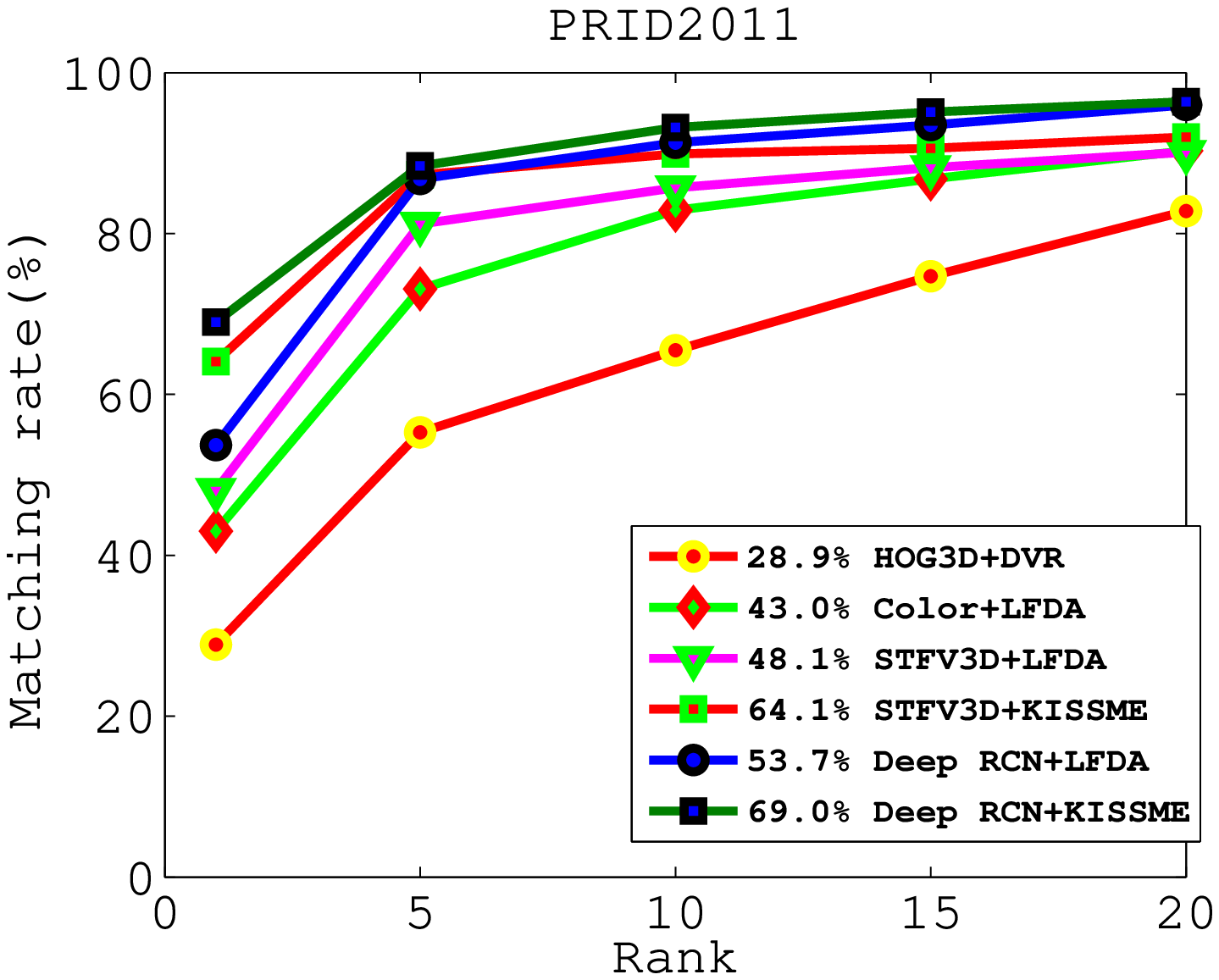}\\
  (b) PRID2011
\caption{CMC curve on the iLIDS-VID and PRID2011 datasets. Rank-1 matching rate is marked before the name of each approach.}
\label{fig:match_rate}
\end{figure}

Table \ref{tab:com_state} and Fig.\ref{fig:match_rate} show the results of the comparison. Distance metric learning can further improve the performance of our appearance descriptor. Notably our method combined with KISSME achieves rank-1 accuracy of $46.4\%$ and $69.0\%$ on iLIDS-VID and PRID2011 datasets, outperforming all benchmarking methods.

\begin{table*}[t]
\caption{Comparison with state-of-the-art methods.}\label{tab:com_state}
\centering
\begin{tabular}{l|c|c|c|c|c|c|c|c}
\hline\hline
Dataset & \multicolumn{4}{c|}{iLIDS-VID} & \multicolumn{4}{c}{PRID2011}\\
\cline{1-9}
Rank @ R & R = 1 & R = 5 & R = 10 & R = 20 & R = 1 & R = 5 & R = 10 & R = 20\\
\hline
GEI+RSVM \cite{GaitRank} & 2.8 & 13.1& 21.3 & 34.5 & - & - & - & -\\
HOG3D+DVR \cite{VideoRanking} & 23.3 & 42.4 & 55.3 & 68.4 & 28.9& 55.3 & 65.5 & 82.8\\
Color+LFDA \cite{Pedagadi2013Local} & 28.0 & 55.3 & 70.6 & 88.0 & 43.0 & 73.1 & 82.9 & 90.3\\
STFV3D \cite{VideoPerson}+LFDA \cite{Pedagadi2013Local} & 38.3 & 70.1 & 83.4 & 90.2 & 48.1 & 81.2 & 85.7 & 90.1\\
STFV3D \cite{VideoPerson}+KISSME \cite{KISSME} & 44.3 & 71.7 & 83.7 & 91.7 & 64.1 & 87.3 & 89.9 & 92.0\\
Deep RCN+LFDA \cite{Pedagadi2013Local} & 43.4 & 75.4 & 89.0 & 95.2 & 53.7  &86.8 & 91.3 & 96.0\\
Deep RCN+KISSME \cite{KISSME} & \textbf{46.1} & \textbf{76.8} & \textbf{89.7} & \textbf{95.6} & \textbf{69.0} & \textbf{88.4} & \textbf{93.2} & \textbf{96.4}\\
\hline
\hline
\end{tabular}
\end{table*}

%% file: conclusion.tex
\section{Conclusion}\label{sec:con}
In this paper, we study the problem of video-based person re-identification and present a deep recurrent network to jointly learn spatio-temporal features and corresponding similarity value given a pair of sequences of images. The proposed approach considers convolutional activations at low levels, which are embeded into recurrent layers to capture motion patterns. Experimental results are evaluated on two standard datasets, and show that our method outperforms state-of-the-arts notable.